\documentclass[10pt, conference, compsocconf]{IEEEtran}
\ifCLASSINFOpdf
\else
\fi

\usepackage{mathrsfs}

\usepackage{array}
\IEEEoverridecommandlockouts
\usepackage{cite}
\usepackage{algorithmic}
\usepackage{times}
\usepackage{epsfig}
\usepackage{graphicx}
\usepackage{amsmath}
\usepackage{amssymb}
\usepackage{amsfonts}
\usepackage{booktabs}
\usepackage{color}
\usepackage{colortbl}
\usepackage{subfig}
\usepackage{flushend}
\usepackage{textcomp}
\usepackage{caption}

\usepackage[utf8]{inputenc}

\usepackage{siunitx}

\usepackage{bm}

\definecolor{cwblue1}{rgb}{0.27,0.427,0.623}
\definecolor{cwblue2}{rgb}{0.286,0.454,0.658}
\definecolor{cwblue3}{rgb}{0.733,0.811,0.905}

\graphicspath{{./Figs/}}

\begin{document}
%
\title{Manifold Mixup improves text recognition with CTC loss}


\author{
\IEEEauthorblockN{Bastien Moysset
\IEEEauthorblockA{A2iA SA, Paris, France}
}
\and
\IEEEauthorblockN{Ronaldo Messina
\IEEEauthorblockA{A2iA SA, Paris, France}
}
}

\maketitle

%

%

\textbf{\textit{Abstract} --- Modern handwritten text recognition techniques employ deep recurrent neural networks. The use of these techniques is especially efficient when a large amount of annotated data is available for parameter estimation. Data augmentation can be used to enhance the performance of the systems when data is scarce. Manifold Mixup is a modern method of data augmentation that meld two images or the feature maps corresponding to these images and the targets are fused accordingly. We propose to apply the Manifold Mixup to text recognition while adapting it to work with a Connectionist Temporal Classification cost. We show that Manifold Mixup improves text recognition results on various languages and datasets. }\\


\section{Introduction}

Text recognition is an important step in most document image analysis applications. It enables to automatically access the information contained in the pages.   

Huge improvement of handwritten text recognition systems has been obtained during the last decade. On the one hand, this amelioration is due to the recognition of text lines using the Connectionist Temporal Classification (CTC) \cite{Graves06icml} in order to implicitly align the image and the target sequence. On the other hand, this is enabled by modern recurrent neural network techniques, whether based on interleaved convolutional and 2D Long Short-Term Memory (LSTM) layers \cite{graves2009offline,moysset2014a2ia} or on convolutions followed by 1D-LSTM layers \cite{puigcerver2017multidimensional,Bluche2017}.

\subsection{State of the Art}

The use of neural networks helped to create systems that can cope with high style heterogeneity within a character class. However, these powerful algorithms with a high number of trained parameters do need a large amount of annotated images to reach an optimal performance. 

Several methods have been proposed to reduce the need of annotated data when training the text recognition systems.

First, in the line of what has been proposed in image classification \cite{krizhevsky2012imagenet}, data augmentation can be performed to enlarge the number of training samples. Real images can be slanted and stretched to form new samples \cite{moysset2014a2ia} or they can be warped with a random grid-based distortion \cite{wigington2017data}.

Secondly, the training set can be extended by adding to it artificial images. This can be done by using handwritten-like fonts \cite{helmers2003generation} or by creating text line images from a recomposition of individually extracted real letter images \cite{elarian2015arabic,shen2016method}. Artificial text images can also be synthetized from a recurrent model trained to estimate the ink paths \cite{Graves13}. More recently, Alonso et al. \cite{alonso2019Gan} proposed a generative adversarial network (GAN) based technique to synthetize handwritten word images. 

On the other hand, the lack of training data should be fought to prevent the network from overfitting. For this, regularization techniques like weight decay or dropout can be applied during the network training \cite{pham2014dropout}.

Recently, Mixup strategies have been proposed in the Machine Learning community, and mainly for object classification tasks, with the aim to cope with reduced amount of available data. The common idea of these strategies is to fuse several (usually two) images or their transformations and use the interpolated resulting image as an input for the training.

\begin{figure}[t]
    \centering
    \includegraphics[width=\columnwidth]{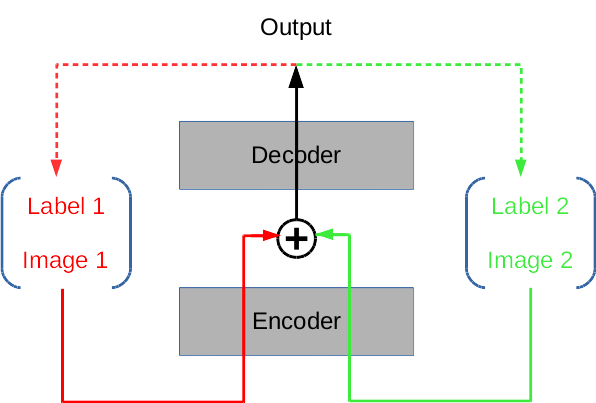}
    \caption{Illustration of the Manifold Mixup principle. Two images, respectively shown in red and green, are mixed after being encoded. The fused features are passed through a decoder and the common output is compared separately to both image labels.}
    \label{fig:1_principle}
\end{figure}

Inspired by Chawla et al. \cite{chawla2002smote} that interpolate input features of objects of the same class, DeVries et al. \cite{devries2017dataset} proposes to mix samples from the same class after being encoded by some of the neural network layers. 
Input images of different classes can also effectively be fused if the cost function is adapted so that the network learns to estimate an interpolation of the labels \cite{zhang2017mixup} and the mixing strategy itself can be learned to avoid manifold intrusions \cite{guo2018mixup}. 

Verma et al. \cite{verma2018manifold} merges both these ideas and propose to mix the labels and images from different classes, or their encoding, at various layers in the network. They show improved results on unseen data and resistance to adversarial examples. 

We based our proposed technique on this method from Verma et al. and will discuss it more in details in Section \ref{sec_manifoldMixup}.  

\subsection{Problem statement}

In this work, we tackle with a manifold mixup based approach illustrated in Figure \ref{fig:1_principle}, the following issues of training neural networks for handwritten text recognition:
\begin{itemize}
\item{The heterogeneity of the handwritten text images to recognize due to varying styles between writers and to different document backgrounds.}
\item{A rather small amount of annotated data available making it harder to generalize on unseen images.}
\item{A trend to overfitting due to both the upper mentioned items and to the potentially high number of parameters in the network.}
\end{itemize}

We make several contributions in order to create a manifold mixup system that cope with the specificities of the text recognition task (compared to image classification). 
\begin{itemize}
\item{We introduce a padding and a width-based grouping strategy within mini-batches in order to handle the varying size of the input images.}
\item{We propose a fusion of gradients from two CTC losses in order to mix the two image targets which are label sequences of unequal lengths. }
\item{We study the impact of the position where the mixup is done and of other implementation choices on a standard recurrent text recognition network.}
\item{We prove the effectiveness of the proposed method on a set of handwritten databases in several languages and with varying sizes.}
\end{itemize}

We will describe in details the manifold mixup strategy in Section \ref{sec_manifoldMixup}. The adaptation to a CTC recognition training will be explained in Section \ref{sec_ctcMixup}. Finally, experimental results will be shown in Section \ref{sec_experimental}.

\section{Manifold Mixup}
\label{sec_manifoldMixup}

\begin{figure}[t]
    \centering
    \includegraphics[width=\columnwidth]{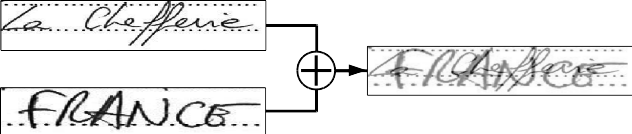}
    \center{(a)}
    \includegraphics[width=\columnwidth]{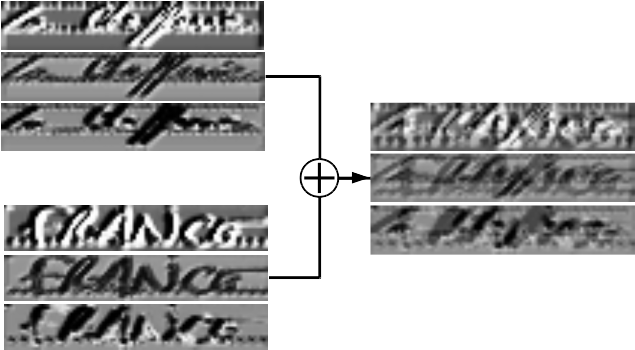}
    \center{(b)}
    \caption{Illustration of the Mixup process. Two feature map sets (left) and their interpolation (right) are shown, both for the input image space (a) and for the space encoded by the 4 first layers of the network (b). }
    \label{fig:mixIllus}
\end{figure}

The Manifold Mixup strategy for training classifiers was introduced by Verma et al. \cite{verma2018manifold}. It is related to the input image Mixup strategy of Zhang et al. \cite{zhang2017mixup}.
The main idea of these methods is to, during training, do a randomly weighted interpolation of two images (or of the transformation of these images) and of their respective labels. The Figure \ref{fig:mixIllus} illustrates this interpolation, both within the input image space and in a feature map space obtained after forwarding the images through the first layers of the network.

The idea behind Mixup training is the following: in the image to labels space, we want to model the manifold that transforms any image into its label. For this, we use (input, label) data that correspond to points in this space. Because the number of trainable parameters is usually higher than the number of training data available, this setup is prone to overfitting. Using Mixup means that we do not only learn from the data points, but from all the segments linking any two data couples. 

More formally, consider a neural network made of $N$ stacked layers. For a layer $k$ of the neural network, let $h_k$ be the function calculated by the layers of the neural network up to this layer $k$. Similarly, let $g_k$ be the function made by the layers from this layer $k$. If $f$ is the function corresponding to the full network, it means that for an input image $\mathbf{x}$, we have:

\begin{equation}
\forall k \in [ 1 , N ] , f(\mathbf{x}) = g_k(h_k(\mathbf{x}))
\end{equation}

The $k$ value is chosen randomly between several possible values. The Mixup function $f_{mixup}$, for a random weighting value $\lambda$, is applied to two training data input-label couples ($\mathbf{x_i},\mathbf{y_i}$) and ($\mathbf{x_j},\mathbf{y_j}$), so that:

\begin{equation}
f_{mixup}(\mathbf{x_i} , \mathbf{x_j}, \lambda) = g_k (\lambda h_k(\mathbf{x_i}) +  (1 - \lambda) h_k(\mathbf{x_j}) )
\end{equation}

The loss $l(\mathbf{\hat{y}}, \mathbf{y})$ we optimize for is a function of an input and a label where the estimated output is $\mathbf{\hat{y}} = f(\mathbf{x})$. For a classification problem with $M$ classes, this loss is a cross entropy:

\begin{equation}
l(\mathbf{\hat{y}}, \mathbf{y}) = - \sum_{c=1}^M y_{c} log(\hat{y_{c}})
\end{equation}

For Mixup, the loss $L_{mixup}$ is computed from the output $\mathbf{\hat{y}}$ of the network and the weighted interpolation of the two labels.  

\begin{equation}
L_{mixup} = l(f_{mixup}(\mathbf{x_i} , \mathbf{x_j} , \lambda), \lambda \mathbf{y_i} + (1 - \lambda) \mathbf{y_j}) 
\label{eq_mixupLoss}
\end{equation}

\begin{equation}
L_{mixup}(\mathbf{\hat{y}}, (\mathbf{y_i}, \mathbf{y_j}) ) = - \sum_{c=1}^M (\lambda y_{i,c} + (1 - \lambda) y_{j,c} ) log(\hat{y_{c}}) 
\end{equation}

We can observe that this loss is the weighted sum of the individual losses: 

\begin{equation}
L_{mixup}(\mathbf{\hat{y}}, (\mathbf{y_i}, \mathbf{y_j}) ) = \lambda l(\mathbf{\hat{y}}, \mathbf{y_i}) + (1-\lambda) l(\mathbf{\hat{y}}, \mathbf{y_j})
\label{eq_classifMixupLoss}
\end{equation}

And that the gradients are such that:

\begin{equation}
\frac{\partial L_{mixup}(\mathbf{\hat{y}}, (\mathbf{y_i}, \mathbf{y_j}) )}{\partial \hat{y_c}} =  \lambda \frac{ \partial l(\mathbf{\hat{y}}, \mathbf{y_i})}{ \partial \hat{y_c}} + (1-\lambda) \frac{ \partial l(\mathbf{\hat{y}}, \mathbf{y_j}) }{\partial \hat{y_c}}
\label{eq_classifDerivative}
\end{equation}

\section{Text line recognition}

\subsection{Gated Convolutional Network}

For the text line recognition problem, we use a Gated Convolutional Network (GNN) inspired by Bluche et al. \cite{Bluche2017}.
The grayscale input images are isotropically rescaled to a fixed height of 128 pixels, normalized and passed through a 2 $\times$ 2 tiling. 
The network is composed of 12 layers, as indicated in Table \ref{tab:netArchi} and in Figure \ref{fig:networkIllustration}. The first 8 layers are convolutional and some of them are used as gates which means that their outputs are pointwise multiplied with their inputs. On top of this convolutional encoder, a vertical max-pooling is applied and a recurrent decoder, comprising two Bidirectional Long Short-Term Memory (LSTM) and two linear layers, is applied to the obtained one dimensional signal.
The output has the depth of the labelset size (including the blank) and a width proportional to the width of the input image. 

\begin{table}[h!]
    \centering
    \begin{tabular}{|l|c|c|c|c|}
        \hline
        Layer & input Depth & output Depth & Filter Size & Stride   \\
        \hline
        Tiling & 1 & 4 & 2 $\times$ 2 & 2 $\times$ 2 \\
        Convolution & 4 & 8 & 3 $\times$ 3 & 1 $\times$ 1 \\
        Convolution & 8 & 16 & 4 $\times$ 2 & 4 $\times$ 2 \\
        Gated Conv. & 16 & 16 & 3 $\times$ 3 & 1 $\times$ 1 \\
        Convolution & 16 & 32 & 3 $\times$ 3 & 1 $\times$ 1 \\
        Gated Conv. & 32 & 32 & 3 $\times$ 3 & 1 $\times$ 1 \\
        Convolution & 32 & 64 & 4 $\times$ 2 & 4 $\times$ 2 \\
        Gated Conv. & 64 & 64 & 3 $\times$ 3 & 1 $\times$ 1 \\
        Convolution & 64 & 128 & 3 $\times$ 3 & 1 $\times$ 1 \\
        Max-Pooling & 128 & 128 & 4 $\times$ 1 & 1 $\times$ 1 \\
        B-LSTM & 128 & 128 & 1 $\times$ 1 & 1 $\times$ 1 \\
        Linear & 128 & 128 & 1 $\times$ 1 & 1 $\times$ 1 \\
        B-LSTM & 128 & 128 & 1 $\times$ 1 & 1 $\times$ 1 \\
        Linear & 128 & 128 & 1 $\times$ 1 & 1 $\times$ 1 \\
        \hline
    \end{tabular}   
    \caption{The architecture of the neural network used for text recognition.}
    \label{tab:netArchi}
\end{table}

\subsection{Training with the CTC}

The text line recognition problem has some specificities in comparison to classification. The output of the network is not a vector of probabilities, but a sequence $\mathbf{\hat{y}} = \{ \mathbf{\hat{y^1}} ... \mathbf{\hat{y^T}} \}$ of vectors of probabilities where the sequence size $T$ is varying and proportional to the width of the text line image. Similarly, the target is not a label $\mathbf{y}$ but a sequence of labels $\mathbf{l}=\{ \mathbf{l_1} ... \mathbf{l_S} \} $ where the label sequence size $S$ corresponds to the number of characters in the text line.

Because $S$ and $T$ are both different and varying independently ($T > S$), we need to perform some sort of alignment between the predictions and the targets. This alignment is implicitly made by the Connectionist Temporal Classification (CTC) \cite{Graves06icml}.

If we define $\mathcal{B}(\mathbf{l})$ as the function that transform a given label sequence $\mathbf{l}$ in all the possible sequences, including blanks, of size $T$, the CTC loss $L_{ctc}$ is defined as: 

\begin{equation}
L_{ctc}(\mathbf{\hat{y}},\mathbf{l}) = -log\left(\sum_{\bm{\pi} \in \mathcal{B}(\mathbf{l})} \prod_{t=1}^T y_{\pi_t}^t\right) 
\end{equation}

This cost can be computed efficiently using dynamic programming alongside the gradients that will be backpropagated in the network.

\begin{equation}
 \frac{ \partial L_{ctc}(\mathbf{\hat{y}}, \mathbf{l}) }{\partial \hat{y_{c}^{t}}} = - \frac{1}{\hat{y_{c}^{t}} \sum_{ \bm{\pi} \in \mathcal{B}(\mathbf{l})} \prod_{z=1}^T y_{\pi_z}^z} \sum_{\bm{\pi} \in \mathcal{B}(\mathbf{l}) | \pi_t = k} \prod_{z=1}^T y_{\pi_z}^z
\end{equation}

\section{Applying Mixup to text recognition}
\label{sec_ctcMixup}

\begin{figure}[t]
    \centering
    \includegraphics[width=0.95\columnwidth]{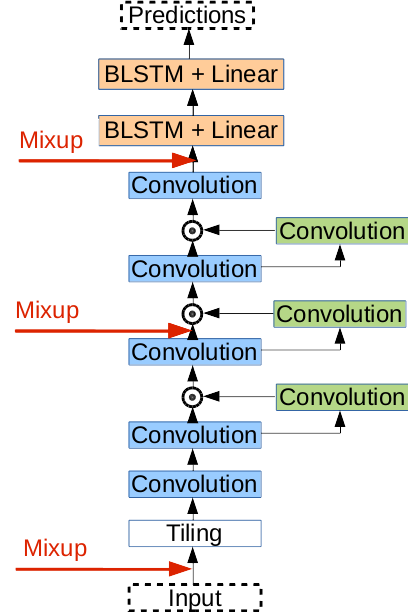}
    \caption{Illustration of the architecture of our text recognition neural network. Standard convolutional layers are shown in blue, while gated convolutional layers are in green and recurrent layers in orange. The red arrows indicate the places in the network where mixup is performed.}
    \label{fig:networkIllustration}
\end{figure}

In this work, we aim at applying a mixup strategy to the sequence recognition problem. It means that we want, at a given level $k$, to fuse the feature maps $h_k(\mathbf{x_i})$ and $h_k(\mathbf{x_j})$ from two text line input images $\mathbf{x_i}$ and $\mathbf{x_j}$. In practice, we choose $k$ randomly within \{0,4,8\} as illustrated in Figure \ref{fig:networkIllustration}.

Because $\mathbf{x_i}$ and $\mathbf{x_j}$ may have different horizontal sizes, we pad (with white) all the images from a mini-batch to the max horizontal size within the mini-batch. In order to save computing time and to get more coherent mixups, we group within a same mini-batch images of similar width in the line of what is done in bucketing techniques \cite{khomenko2016accelerating}. 
Images are weighted with a random valued $\lambda$ ratio. In practice, $\lambda$ is drawn from a $Beta(0.5)$ distribution. After going through the end of the network, we obtain a common sequence of prediction probabilities for both lines. 

Nevertheless, interpolating the targets like it is done in equation \ref{eq_mixupLoss} is not possible due to the fact that the two label sequences have different lengths and because the forward-backward algorithm used to compute the CTC works well only for one-hot encodings of targets. 

For this reason we choose to directly use as a loss $L_{ctc\_mixup}$ the weighted sum of two CTC losses, one with each label sequence $\mathbf{l_i}$ and $\mathbf{l_j}$.

\begin{equation}
L_{ctc\_mixup}(\mathbf{\hat{y}}, (\mathbf{l_i}, \mathbf{l_j})) = \lambda L_{ctc}(\mathbf{\hat{y}}, \mathbf{l_i}) + (1-\lambda) L_{ctc}(\mathbf{\hat{y}}, \mathbf{l_j})
\end{equation}

We note that this weighting of losses is analogous to what was observed for the classification in Equation \ref{eq_classifMixupLoss} and, like in Equation \ref{eq_classifDerivative}, we get gradients relatively to the weighted gradients from the two targets.

\begin{equation}
\frac{\partial L_{ctc\_mixup}(\mathbf{\hat{y}}, (\mathbf{y_i}, \mathbf{y_j}) )}{\partial \hat{y_{c}^{t}}} =  \lambda \frac{ \partial L_{ctc}(\mathbf{\hat{y}}, \mathbf{y_i})}{ \partial \hat{y_{c}^{t}}} + (1-\lambda) \frac{ \partial L_{ctc}(\mathbf{\hat{y}}, \mathbf{y_j}) }{\partial \hat{y_{c}^{t}}}
\end{equation}

\section{Experiments}
\label{sec_experimental}

\subsection{Experimental setup}
\label{sec_expeSetup}

We performed most of the experiments on the handwritten lines in French from the Maurdor dataset \cite{maurdor}. This dataset has the particularity of being very challenging with heterogeneous images from different kind of documents (forms, letters, drawings, ...) and various scanning procedures.
We also made control experiments on other handwritten sub-datasets from Maurdor (English and Arabic), on the easier tasks RIMES \cite{rimes} and IAM \cite{iam} and on the Chinese CASIA \cite{liu2011casia} dataset (we do not use the isolated characters parts). The statistics regarding the number of lines in the used datasets are available in Table \ref{tab:expeSetupStats}.

\begin{table}[h!]
    \centering
    \begin{tabular}{|l|c|c|}
        \hline
        Dataset & Training lines & Validation lines \\
        \hline
        Maurdor French Handwritten & 26870 & 2054 \\
        Maurdor English Handwritten & 10825 & 1115 \\
        Maurdor Arabic Handwritten & 11905 & 1125 \\
        RIMES & 10532 & 801 \\
        IAM & 6482 & 976 \\
        CASIA & 35856 & 5914 \\
        \hline
    \end{tabular}   
    \caption{Number of lines present in the datasets used in this work.}
    \label{tab:expeSetupStats}
\end{table}

To assess the performances of the models, we compute a character error rate (CER) as the Levenshtein distance between the predicted and the ground truth sequences.
Excepted for the Chinese, where a language model is used to recover the characters from an encoding as detailed in Bluche et al. \cite{BlucheM16}, we do not use any language model and only the agglutinated best predictions are considered.

For all the models, we use a Glorot \cite{glorot2010understanding} initialization of the weights, RMSProp \cite{rmsprop} based gradient descent, mini-batches of size 8, and a learning rate of $4.10^{-4}$. Models are selected with an early stopping on the validation set after 200 epochs without improvement.

\subsection{Ablation study}

In this section, we validate and discuss the key design choices we made for our system.  

\subsubsection{Impact of the position of the Mixup}

First, we compare, in Table \ref{tab:expe_mixupPosition}, the performance of our handwriting recognition system with respect to the position where the mixup of the feature maps is performed. We observe that, contrarily to what was observed by Zhang et al. \cite{zhang2017mixup} for several tasks, the performances with input mixup are worse than the baseline (no mixup) model. Accordingly to this same paper \cite{zhang2017mixup}, our recognition performance decreases if the mixup is performed in higher latent spaces.

However, we observe that performing the mixup randomly at one of several positions in the network, like what was done in Verma et al. \cite{verma2018manifold}, helps to improve the results compared to the no-mixup strategy. This is probably due to the fact that by adding some randomness, we force the network to learn with the interpolations and prevent it from disentangling the signals. No improvement was obtained by training both with and without mixup.

\begin{table}[h!]
    \centering
    \begin{tabular}{|l|c|}
        \hline
          & CER (\%)    \\
        \hline
        No mixup & 9.39  \\
        At the input & 9.55 \\
        After the 4th convolutional layer & 10.10   \\
        After the 8th convolutional layer & 11.60   \\
        Randomly at one of the 3 positions & 8.91 \\
        Randomly at one of the 3 positions or no fusion & 9.15 \\
        \hline
    \end{tabular}   
    \caption{Comparison of the CER with respect to the position(s) in the network where the feature maps from different images are fused.}
    \label{tab:expe_mixupPosition}
\end{table}

\subsubsection{Number of images mixed-up}

As shown in Table \ref{tab:expe_numMixedImages}, we do not observe further improvement of the results when mixing 3 images instead of 2.

\begin{table}[h!]
    \centering
    \begin{tabular}{|l|c|}
        \hline
          & CER (\%)    \\
        \hline
        Fusion of 2 images & 8.91  \\
        Fusion of 3 images & 9.02  \\
        \hline
    \end{tabular}   
    \caption{Comparison of CER when fusing 2 or 3 image feature maps.}
    \label{tab:expe_numMixedImages}
\end{table}

\subsubsection{Multiplication of the gradients by the Mixup ratio}

A key component of our system is the choice to multiply the gradients (or the loss, this is the same) by the same $\lambda$ random value that is used to weight the two feature maps during the mixup. To confirm the validity of this choice, we train the network without this multiplication of the gradients by $\lambda$. It means that the network is trained to recognize both label sequences without taking into account the weighting ratio. Results, found in Table \ref{tab:expe_gradMultiplications}, show that doing the multiplication by $\lambda$ is indeed very important.

\begin{table}[h!]
    \centering
    \begin{tabular}{|l|c|}
        \hline
          & CER (\%)    \\
        \hline
        With gradient multiplication  & 8.91  \\
        Without gradient multiplication  & 10.08   \\
        \hline
    \end{tabular}   
    \caption{Influence on the CER of the multiplication of gradients by the mixup ratio.}
    \label{tab:expe_gradMultiplications}
\end{table}

\subsubsection{Impact of the mixup ratio distribution.}

This $\lambda$ parameter is drawn from a $Beta(0.5)$ distribution, as in Zhang et al. \cite{zhang2017mixup}. Verma et al. \cite{verma2018manifold} use a $Beta(2)$ distribution. In Table \ref{tab:expe_distributionRatios}, we compare the results with both these distributions alongside uniform distributions. No significative differences can be observed between the distributions. 

\begin{table}[h!]
    \centering
    \begin{tabular}{|l|c|}
        \hline
        Distribution    & CER (\%)  \\
        \hline
        Uniform [0,1] (=$Beta(1)$) & 8.92   \\
        Uniform [0.1 , 0.9] & 8.95   \\
        $Beta(0.5)$ & 8.91 \\
        $Beta(2)$ & 9.12 \\
        \hline
    \end{tabular}   
    \caption{Impact of the chosen mixup ratio $\lambda$ distribution on the CER.}
    \label{tab:expe_distributionRatios}
\end{table}

\subsection{Analysis of Mixup results}

In order to prove the robustness of the presented manifold mixup approach, we compare the performances on the list of datasets introduced in Section \ref{sec_expeSetup}. Results can be found in Table \ref{tab:expe_results}. We observe a consistent decrease of the character error rates on the difficult datasets that are the three datasets from Maurdor and on the CASIA dataset. Results on the more simple RIMES and IAM datasets are similar with and without mixup.

\begin{table}[h!]
    \centering
    \begin{tabular}{|l|c|c|}
        \hline
        Dataset     & Without mixup   & With mixup  \\
        \hline
        Maurdor French Handwritten & 9.39 & 8.91 \\
        Maurdor English Handwritten & 16.0 & 14.8 \\
        Maurdor Arabic Handwritten & 11.0 & 10.5 \\
        CASIA & 27.5 & 23.9 \\
        RIMES & 3.30 & 3.32 \\
        IAM & 4.68 & 4.64 \\
        \hline
    \end{tabular}   
    \caption{Impact of the use of feature map mixup during the training on the CER (\%), for various datasets.}
    \label{tab:expe_results}
\end{table}

The significativity of the improvement observed when adding the manifold mixup is addressed in Table \ref{tab:expe_reproducibility} by training 10 different networks (meaning that 10 different random seeds are used for initialization), both with and without Mixup, on the Maurdor French Handwritten dataset. 

\begin{table}[h!]
    \centering
    \begin{tabular}{|l|c|c|c|c|}
        \hline
            & Min  & Max  & Mean & Median  \\
        \hline
        Without mixup  & 9.08 & 9.46 & 9.30 & 9.32 \\
        With mixup  & 8.65 & 9.11 & 8.88 & 8.91 \\
        \hline
    \end{tabular}   
    \caption{Study of the reproducibility over 10 trainings with and without mixup. CER (\%) are shown.}\label{tab:expe_reproducibility}
\end{table}

We also compare, in Table \ref{tab:expe_dataSize}, the impact of the amount of training data available on the performances. We observe that the relative improvement is progressively increasing from 5.1\% to 12.3\% when the dataset size is reduced from 26 000 to 5 000 examples. This illustrates how using manifold mixup for text recognition does increase the generalization ability of the network.

\begin{table}[h!]
    \centering
    \begin{tabular}{|c|c|c|}
        \hline
        Number of images  & With Mixup & Without Mixup  \\
        \hline
        26870 (All) & 8.91 & 9.39  \\
        20000 & 9.85 & 10.4 \\
        10000 & 13.3 & 14.4  \\
        5000 & 20.6 & 23.5  \\
        \hline
    \end{tabular}
    \caption{Impact of the number of training examples available on the CER (\%), on the Maurdor French Handwritten set.}\label{tab:expe_dataSize}
\end{table}

Similarly, we observe that using mixup strongly diminishes the overfitting. In particular, we observe that the CTC loss on the validation set tends to go up at some point without manifold mixup, while it reaches some sort of horizontal asymptote when mixup is performed. This confirms that the mixup approach has some regularization effect on the network as mentioned in previous works. \cite{zhang2017mixup,guo2018mixup,verma2018manifold}.

Following this line of reasoning, we study, in Table \ref{tab:expe_regularization} the cross impact of an other common layer with a regularization effect, the Dropout \cite{srivastava2014dropout}, that is used in our training process. We can see that both of them, independently strongly improve the results of the text recognition. But their improvements are not completely co-linear as using both of them further improve the performances.

\begin{table}[h!]
    \centering
    \begin{tabular}{|c|c|c|}
        \hline
        Dropout  & Manifold Mixup  & CER (\%)  \\
        \hline
        No & No &  15.4  \\
        Yes & No & 9.39 \\
        No & Yes & 10.6  \\
        Yes & Yes & 8.91  \\
        \hline
    \end{tabular}
    \caption{Cross study of two regularizer, dropout and manifold mixup, on Maurdor French Handwritten text recognition results (CER).}
    \label{tab:expe_regularization}
\end{table}

%
%

\section{Conclusion}

In this paper, we presented a new training strategy for text recognition systems based on Manifold Mixup. We proved that this technique acts as a strong regularizer by interpolating the input images and the gradients. We proposed a technique to apply the mixup to varying size images and sequence prediction with Connectionist Temporal Classification alignments.
We demonstrated a significative improvement of text recognition results on several handwritten datasets of varying sizes and languages. \\
\\


\vspace{-0.5cm}

\bibliographystyle{IEEEtran}
\bibliography{IEEEabrv,mixup}

\end{document}